\documentclass[conference]{IEEEtran}
\IEEEoverridecommandlockouts
\usepackage{cite}
\usepackage{amsmath,amssymb,amsfonts}
\usepackage{algorithmic}
\usepackage{graphicx}
\usepackage{textcomp}
\usepackage{xcolor}
\usepackage{multirow}
\usepackage{amsthm}
\usepackage{statmath}
\usepackage{epsfig}
\usepackage{pdfpages}
\usepackage{bm} 
\usepackage[hidelinks]{hyperref}

\DeclareMathOperator{\EX}{\mathbb{E}}

\font\myfont=cmr12 at 19pt

\begin{document}
\bstctlcite{IEEEexample:BSTcontrol}
\title{\myfont Learning Generalized Hybrid Proximity Representation for Image Recognition}

\author{\IEEEauthorblockN{1\textsuperscript{st} Zhiyuan Li \thanks{Copyright (c) 2022 IEEE. Personal use of this material is permitted.
 Permission from IEEE must be obtained for all other uses, in any current or future 
 media, including reprinting/republishing this material for advertising or promotional 
 purposes, creating new collective works, for resale or redistribution to servers or 
 lists, or reuse of any copyrighted component of this work in other works.}
  \thanks{Contact authors: li3z3@mail.uc.edu (Zhiyuan Li), ralescal@ucmail.uc.edu (Anca Ralescu).}}
\IEEEauthorblockA{\textit{Department of Computer Science} \\
\textit{University of Cincinnati}\\
Cincinnati, OH, United States \\
li3z3@mail.uc.edu}
\and
\IEEEauthorblockN{2\textsuperscript{nd} Anca Ralescu}
\IEEEauthorblockA{\textit{Department of Computer Science} \\
\textit{University of Cincinnati}\\
Cincinnati, OH, United States \\
ralescal@ucmail.uc.edu}
}

\maketitle

\begin{abstract}
Recently, deep metric learning techniques received attentions, as the learned distance representations are useful to capture the similarity relationship among samples and further improve the performance of various of supervised or unsupervised learning tasks. We propose a novel supervised metric learning method that can learn the distance metrics in both geometric and probabilistic space for image recognition. In contrast to the previous metric learning methods which usually focus on learning the distance metrics in Euclidean space, our proposed method is able to learn better distance representation in a hybrid approach. To achieve this, we proposed a \textbf{\emph{Generalized Hybrid Metric Loss}} (GHM-Loss) to learn the general hybrid proximity features from the image data by controlling the trade-off between geometric proximity and probabilistic proximity. To evaluate the effectiveness of our method, we first provide theoretical derivations and proofs of the proposed loss function, then we perform extensive experiments on two public datasets to show the advantage of our method compared to other state-of-the-art metric learning methods. 
\end{abstract}


\section{Introduction}
Metric learning takes input data to learn the similar and dissimilar features between samples. The learned distance metric provides a meaningful and robust representation to discriminate the proximity or distance between samples and can be further utilized for both supervised and unsupervised learning tasks \cite{kulis2013metric}. Recently, deep learning-based metric learning algorithms, i.e., deep metric learning, were widely applied in the computer vision area by developing either a novel network architecture or an intuitive and efficient loss function \cite{hoffer2015deep,wang2017deep,kaya2019deep}. Some typical works, such as the Siamese network \cite{koch2015siamese}, Triplet network \cite{hoffer2015deep}, SupCon \cite{khosla2020supervised}, aim to formulate an instance discrimination task to learn a useful feature representation by optimizing the proximity function in the Euclidean space, i.e., geometric distance or Cosine proximity between the feature embeddings. In this paper, we seek to address  the inadequacies of geometric proximity of recent state-of-the-art metric learning methods by reconsidering an alternative approach in which learned distance metrics are not biased to only geometric proximity. 

Metric learning methods have shown an excellent classification performance in image recognition applications, due to their extraordinary ability to discriminate similar information between samples \cite{yang2006distance,davis2007information,kulis2013metric,kaya2019deep}. Such metric learning tasks can be either supervised or self-supervised. In supervised metric learning, the model learns to pull together the samples from the same classes and push away the samples from different classes \cite{le2020contrastive}. Self-supervised metric learning, also named contrastive learning, requires a data augmentation step to create some pseudo-ground-truth from the data itself, where the augmentation from the same sample is included in “positive” pairs and the augmentation from different samples is included in “negative” pairs \cite{jaiswal2020survey}. Similar to supervised metric learning, the self-supervised model learns similar representations from the positive pairs and should be different than the representations of the negative pairs. Various types of metric/contrastive learning works have been developed for image pattern recognition applications, including image classification \cite{wang2021contrastive,khosla2020supervised,he2020momentum,chen2020simple}, image clustering \cite{do2021clustering,zhong2021graph}, image segmentation \cite{chaitanya2020contrastive,hu2021region}, image reconstruction \cite{chen2022unpaired,zheng2021weakly}, and object detection \cite{kim2022spatial}. All these works used geometric proximity (e.g., Cosine similarity or Euclidean distance) as the proximity function in the training an objective loss to learn the geometric representation of the samples. However, the probability distribution of the samples should not be ignored. 

\begin{figure}[ht]
    \centering
    \includegraphics[width=7.5cm]{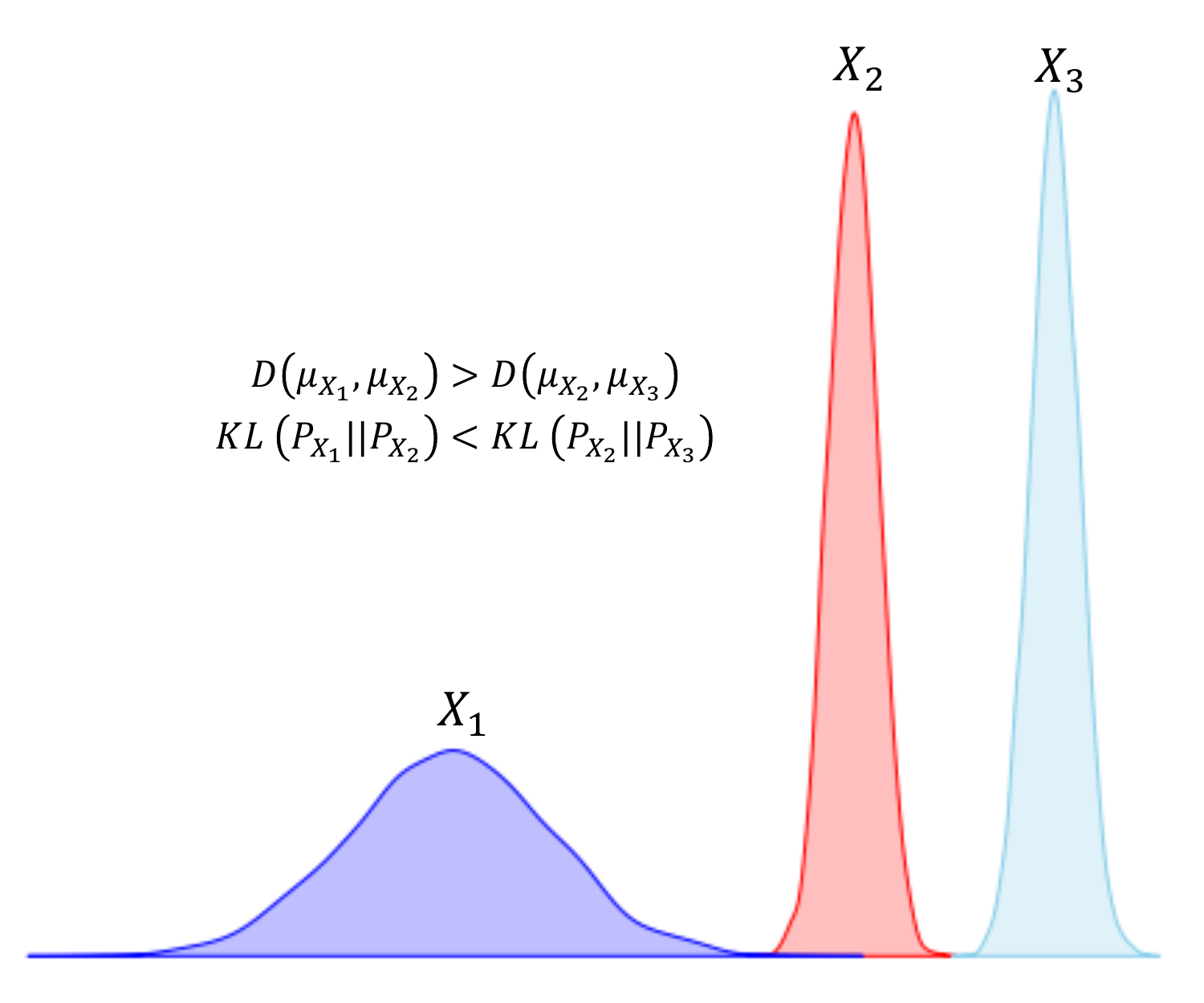}
    \caption{Geometric distance vs. probabilistic distance between probability distributions $X_1,X_2$ and $X_3$. Without considering the variances, the geometric distance between mean values of $\mu_{X_1}, \mu_{X_2}$, and $\mu_{X_3}$ cannot represent their probabilistic distance. \emph{D}: Geometric distance; \emph{KL}: Kullback–Leibler divergence.}
    \label{f1}
\end{figure}

To overcome these limitations and boost the prediction performance of metric learning, we proposed a novel supervised metric learning method to learn the hybrid proximity that combines the proximity in both geometric and probabilistic space. To achieve it, we defined a supervised \textbf{\emph{Generalized Hybrid Metric Loss}} (GHM-Loss) to better learn the distance representations in both geometric and probabilistic space. We noticed that even if the geometric distance is small, the probabilistic distance can be large when the sample variance is large (\textbf{Figure \ref{f1}}). This observation reminds us to reconsider that the model may not sufficiently learn the distance features based only on the geometric distance between data points. Thus, enabling the model to partially learn the probabilistic distance controls the trade-off between two types of distance representation (geometric and probabilistic).

Our proposed GHM-Loss is formulated by a hybrid divergence underlying the geometric and probabilistic space, which is a generalized distance loss form for many defined metric learning methods, including Triplet \cite{hoffer2015deep}, N-pairs \cite{sohn2016improved}, Max Margin \cite{wu2017sampling}, NTXent \cite{chen2020simple}, SupCon \cite{khosla2020supervised}, etc. We first theoretically showing the advantage of using the probabilistic distance in metric learning compared to the geometric-based distance, and we employed two public datasets to show the effectiveness of our method compared to other state-of-the-art metric/contrastive learning methods. We also investigated the superiority of the proposed GHM-Loss with other metric/contrastive learning loss functions. To sum up, our main findings and contributions to this work are as follows: 

\begin{enumerate}
\item We propose a novel supervised metric learning method for enhancing the performance of image recognition by defining a \textbf{\emph{Generalized Hybrid Metric Loss}} (GHM-Loss). The proposed GHM-Loss is able to learn better distance representation that controls the trade-off between the geometric-based and the probabilistic distance from feature embeddings. 

\item We define two proximity functions with certain properties in geometric and probabilistic space, respectively, and provide proof for each property. Meanwhile, we theoretically show the advantage of the GHM-Loss by including the probabilistic proximity for learning the distance between distributions.

\item Our approach is supported both by a theoretical discussion and by extensive experiments performed on two common image classification tasks to demonstrate the effectiveness of our method compared to other state-of-the-art metric learning methods.  
\end{enumerate}

\section{Related Work}
In this section, we first discuss some state-of-the-art methods of deep metric learning and some of its applications in the computer vision domain. We further review related works on contrastive learning.

\subsection{Metric Learning}
\subsubsection{Traditional Metric Learning}
The early stage of the machine learning techniques requires a hand-crafted processing step, i.e., feature engineering, such as feature selection and feature extraction before training a machine learning model for supervised (e.g., classification) or unsupervised learning (e.g., clustering) tasks \cite{xing2002distance,yang2006distance,weinberger2009distance}. These methods, including linear projections, i.e., principal component analysis (PCA) \cite{wold1987principal}, decomposition, i.e., non-negative matrix factorization (NMF) \cite{paatero1994positive} to extract useful feature information and are not directly within the classification structure, resulting in a limited performance on the certain complex structure, such as high-dimensional data and non-linearity. Unlike traditional machine learning approaches, metric learning performs the learning process on the data to learn a distance feature representation by decreasing the distance between similar samples and increasing the distance between dissimilar ones in a embedding space. The learned distance features will have a high ability to discriminate the classes of the sample data. Usually, metric learning approaches apply linear transformation techniques to the input data and map it to a new feature space with a higher-class separation \cite{yang2007overview}. However, these methods lack the generalization capability and nonlinear knowledge of the attributes \cite{hu2014discriminative}. 

\subsubsection{Deep Metric Learning}
Unlike traditional metric learning methods, deep metric learning relies on training deep neural networks with activation functions that capture nonlinear properties \cite{kaya2019deep}, and it has dominated metric representation learning in the image recognition community \cite{koch2015siamese,hoffer2015deep,wang2017deep,khosla2020supervised,dong2021deep,sundgaard2021deep,zhou2022enhancing}. For example, the Siamese network \cite{koch2015siamese} used two identical convolutional neural networks (CNNs) to encode a pair of input samples and minimize the contrastive loss to learn the representative distance features. Similar to the Siamese network, Hoffer et al \cite{hoffer2015deep} proposed a Triplet network, including the anchor, positive (similar), and negative (dissimilar) sample, which learns the inequality that the positive sample stays closer to the anchor compared to the negative sample. Afterward, Wang et al \cite{wang2017deep} defined a new Angular loss to constrain the angle at the negative sample from the Triplet network. Later, Sohn \cite{sohn2016improved} proposed an N-pair loss to address the slow convergence problem of the Triplet loss. More recently, Khosla et al \cite{khosla2020supervised} developed a supervised contrastive learning framework with the more general form of metric learning loss, i.e., SupCon, and showed the effectiveness of classification performance compared to the Triplet loss and the N-pair loss. These existing works have shown great promise for metric and feature representation learning in a variety of image classification tasks. Nevertheless, to the best of our knowledge, most previous studies are focused on learning the geometric-based metrics of the embedding space, while the probabilistic-based metrics are usually ignored. In this work, our method is able to learn the meaningful metrics of both geometric and probabilistic space.

\subsection{Contrastive Representation Learning}
The main purpose of deep metric learning and contrastive learning is to train a deep learning model to learn the distance feature representations in an embedding space. The main difference between these two methods is that contrastive learning is closely related to the self-supervised learning domain, which contains a data augmentation step to generalize an arbitrary number of positive and negative sample pairs from each sample \cite{dai2021adaptive}. Given the stunning achievement of self-supervised representation learning, many contrastive learning methods have been developed for various computer vision tasks \cite{chuang2020debiased,park2020contrastive,kang2020contragan,li2021rotation}. For example, Ye et al \cite{ye2019unsupervised} proposed an embedding contrastive learning method with the Siamese network to learn the invariant features of embedding space. Chen et al \cite{chen2020simple} developed a famous contrastive learning framework, SimCLR, in which the model is pretrained to discriminate the positive pairs of data augmentation from the same source image, demonstrating superior performance in ImageNet classification. Similarly, He et al \cite{he2020momentum} proposed the Moco v1, to maximize the proximity between the positive pairs based on the monument network encoder. Additional studies that are similar to SimCLR and Moco, including BYOL \cite{grill2020bootstrap}, SimSam \cite{chen2021exploring} Barlow Twins \cite{zbontar2021barlow}, etc., show the exceptional performance of learned feature representation for further supervised or unsupervised tasks.

\begin{figure}[ht]
    \centering
    \includegraphics[width=8.87cm]{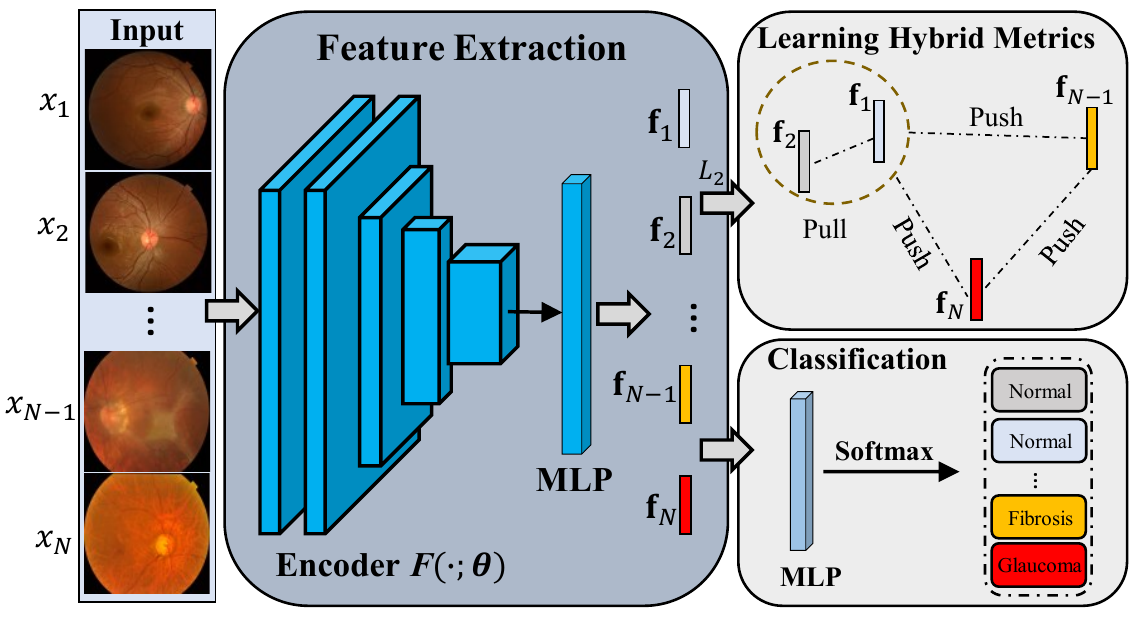}
    \caption{The overview of our proposed framework (example of fundus disease diagnosis). We use a pretrained convolutional neural network (CNN) and a multi-layer perceptron (MLP) to encode each image to the embedded feature. Afterward, we propose a metric learning branch that is supervised with the proposed GHM-Loss which is trained together with the cross-entropy loss of a classification branch in a multi-task scheme.}
    \label{framework}
\end{figure}

\section{Methodology}
\subsection{Overview}
Our proposed supervised metric learning framework is illustrated in \textbf{Figure \ref{framework}}. We first denote a training image dataset $D=\{x_i,y_i\}_{i=1}^{N}$, where $y_i$ is the label of image $x_i$, a set of indices of all the positive samples for a randomly selected image $x_i$ in a batch, $\bm{U}(i)=\{j\in \Theta|y_j=y_i,j\neq i\}$, a set of indices of all negative samples for a randomly selected image $x_i$ in a batch, $\bm{V}(i):=\{j\in \Theta|y_j\neq y_i,i\neq j\}$. The problem formulation is to learn a network $F(\cdot;\theta)$ that maps each input $x_i$ to a $L_2$ normalized $d$-dimensional feature embedding $\textbf{f}_i$, such as $\textbf{f}_i=F(x_i;\theta) \in \mathbb{R}^{d}$. To achieve this, we use a pre-trained CNN, i.e., ResNet18, followed by a MLP to produce $N$ high-level feature vectors and perform two supervised learning branches. The first branch is a metric learning task, which aims to learn the robust metrics by pulling all the samples with indices $\bm{U}(i)$ and pushing away all the samples with indices $\bm{V}(i)$. Meanwhile, the embedded feature $\textbf{f}_i$ is connected to another MLP layer with a Softmax to generate the predicted probability for the class label $y_i$ and is supervised with cross-entropy loss to perform a classification task. Below, we will elaborate on the procedure of the each branch, including the definition of GHM-Loss and its advantage, and other network details.

\subsection{Generalized Hybird Metric Loss}
\subsubsection{General Loss Form}
To perform the metric learning branch, we propose a general form metric loss function, in which the network can learn the proximity information between the embedded features $\{\textbf{f}_1, \textbf{f}_2,\dots, \textbf{f}_N\}$ by optimizing the loss. Let $S(\cdot)$ denote the proximity function for two input vectors $\textbf{f}_{i}$ and $\textbf{f}_{j}$. That is, for $\textbf{f}_{i}, \textbf{f}_{j} \in \mathbb{R}^{d}$, $S(\textbf{f}_{i},\textbf{f}_{j}): \mathbb{R}^d \to \mathbb{R}^1$. Thus, the probability of $x_i,x_u,u\in \bm{U}(i)$ is being recognized as $y_i$ is defined by 
\begin{align}
    p(y_i|x_i,x_u)=\frac{\exp\left[S(\textbf{f}_i,\textbf{f}_u)\right]}{\sum_{j\in \Theta, j\neq i}\exp\left[S(\textbf{f}_i,\textbf{f}_j)\right]} 
    \label{eq1}
\end{align}
On the other hand, the probability of $x_i,x_v,v\in \bm{V}(i)$ is not being recognized as $y_i$ is defined by 
\begin{align}
    p(y_i|x_i,x_v)=\frac{\exp\left[S(\textbf{f}_i,\textbf{f}_v)\right]}{\sum_{j\in \Theta, j\neq i}\exp\left[S(\textbf{f}_i,\textbf{f}_j)\right]} 
    \label{eq2}
\end{align}
Next, assume that all the probabilities of different images being recognized as image $x_i$ are independent, let $q(y_i|x_i,x_v)=1-p(y_i|x_i,x_v)$ thus, the objective likelihood function that we are interested is defined by
\begin{align}
    \ell_i=\prod_{u\in \bm{U}(i)}\prod_{v\in \bm{V}(i)}p(y_i|x_i,x_u)q(y_i|x_i,x_v)
    \label{eq3}
\end{align}
Correspondingly, the  negative log likelihood over all the data points indexed by $\Theta$ yields:
\begin{align}
     \mathcal{L}^*=-\sum_{i\in \Theta}\|\bm{V}(i)\|&\sum_{u\in \bm{U}(i)}\log p(y_i|x_i,x_u) \nonumber \\
     &-\sum_{i\in \Theta}\|\bm{U}(i)\|\sum_{v\in \bm{V}(i)}\log q(y_i|x_i,x_v)
    \label{eq4}
\end{align}
where $\|\bm{U}(i)\|$ and $\|\bm{V}(i)\|$ denotes the size of the set $\bm{U}(i)$ and $\bm{V}(i)$, respectively. 

\subsubsection{Geometric Proximity}
We first consider the proximity in the geometric space. Given a pair vectors $\textbf{f}_{i}$ and $\textbf{f}_{j}$, the proximity function $S_g(\textbf{f}_{i},\textbf{f}_{j})$ satisfies the following properties: 
\begin{itemize}
  \item[1]  $S_g(\textbf{f}_{i},\textbf{f}_{j})\in [0,1]$;
  \item[2]  $S_g(\textbf{f}_{i},\textbf{f}_{j})=S_g(\textbf{f}_{j},\textbf{f}_{i})$;
  \item[3]  $\forall \bm{c}\in [\textbf{f}_{i},\textbf{f}_{j}], S_g(\textbf{f}_{i},\textbf{f}_{j})\le \text{min}\{S_g(\textbf{f}_{i},\bm{c}),S_g(\bm{c},\textbf{f}_{j})\}$.
\end{itemize}
The proximity measures that satisfy the above properties include Cosine similarity. 
\begin{proof}
Using Cosine similarity as the proximity metrics, such that $S_g(\textbf{f}_{i},\textbf{f}_{j})=\textbf{f}_{i}\cdot \textbf{f}_{j}/\|\textbf{f}_{i}\|\|\textbf{f}_{j}\|$ satisfies each property above.
\begin{itemize}
  \item[1.]  Obviously, $S_g(\textbf{f}_{i},\textbf{f}_{j})\in [0,1]$.
  \item[2.]  Obviously, this property is true.
  \item[3.]  Let $\bm{c}\in [\textbf{f}_{i},\textbf{f}_{j}]$, we have $|\textbf{f}_{i}-\bm{c}|\le |\textbf{f}_{i}-\textbf{f}_{j}|$, $|\bm{c}-\textbf{f}_{j}|\le |\textbf{f}_{i}-\textbf{f}_{j}|$, which means that $S_g(\textbf{f}_{i},\bm{c})\ge S_g(\bm{a,b})$, $S_g(\bm{c},\textbf{f}_{j})\ge S(\bm{a,b})$. Thus, $S_g(\textbf{f}_{i},\textbf{f}_{j})\le \text{min}\{S_g(\textbf{f}_{i},\bm{c}),S_g(\bm{c},\textbf{f}_{j})\} \forall \bm{c}\in [\textbf{f}_{i},\textbf{f}_{j}]$.
\end{itemize}
\end{proof}

\subsubsection{Probabilistic Proximity}
Instead of only using geometric proximity, which ignores the sampling probability distribution, we consider the probabilistic proximity to summarize the distribution of the embedded features $\{\textbf{f}_i, \textbf{f}_2, \dots, \textbf{f}_N\}$. Given a pair vectors $\textbf{f}_{i}$ and $\textbf{f}_{j}$, with size of $|\textbf{f}_{i}|,|\textbf{f}_{j}|$, the probabilistic proximity function satisfies the following properties: 
\begin{itemize}
  \item[1.]  $S_p(\textbf{f}_{i},\textbf{f}_{j})\in [0,1]$;
  \item[2.]  $S_p(\textbf{f}_{i},\textbf{f}_{j})=S_p(\textbf{f}_{j},\textbf{f}_{i})$;
  \item[3.]  $S_p(\textbf{f}_{i},\textbf{f}_{j})=0$ \emph{if and only} if $\textbf{f}_{i}=\textbf{f}_{j}$;
  \item[4.]  $S_p(\textbf{f}_{i},\textbf{f}_{j})\le S_p(\textbf{f}_{i},\textbf{f}_c) + S_p(\textbf{f}_c,\textbf{f}_{j})$ under the certain condition, in which $|\textbf{f}_c|=|\textbf{f}_{i}|=|\textbf{f}_{j}|$.
\end{itemize}
We use a Gaussian mixture model (GMM) to represent the empirical distribution $\textbf{f}_i$, which is defined by
\begin{align}
    p(\textbf{f}_{i})=\sum_{k \in K}w_{k}\mathcal{N}(\textbf{f}_i;\mu_{k},\sigma^{2}_{k})
    \label{eq5}
\end{align}
where $w_k$ is a latent variable followed by a categorical distribution, denoting the $k$-th component, and $\mathcal{N}$ is the Gaussian probability density function with parameters $\mu_{k}$ and $\sigma_{k}$, which is defined as 
\begin{align}
    \mathcal{N}(\textbf{f}_i;\mu_{k},\sigma^{2}_{k})=\frac{1}{\sqrt{2\pi\sigma^2_{k}}}\exp{\left[-\frac{1}{2\sigma^2_{k}}(\textbf{f}_i-\mu_{k})^2 \right]} \label{eq6}
\end{align}
Using this model, the probabilistic distance between $p(\textbf{f}_i)$ and $p(\textbf{f}_j)$ is chosen with the symmetric divergence, i.e., Jensen–Shannon (JS)-divergence. For simplicity, we use $\bm{p}_i$ and $\bm{p}_j$ to denotes the probability distributions of $\textbf{f}_i$ and $\textbf{f}_j$, respectively. Therefore, the $S_p(\textbf{f}_{i},\textbf{f}_{j})$ is denoted by
\begin{align}
    S_p(\textbf{f}_{i},\textbf{f}_{j})=\frac{1}{2}\left[d_{\text{KL}}(\bm{p}_i\|\Bar{\bm{p}}_{ij})+d_{\text{KL}}(\bm{p}_j\|\Bar{\bm{p}}_{ij}) \right] \label{eq7}
\end{align}
where $\Bar{\bm{p}}_{ij}=\left(\bm{p}_i+\bm{p}_j\right)/2$, $d_{\text{KL}}(\cdot)$ presents a function of the Kullback–Leibler (KL)-divergence.

\begin{proof}
We prove that $S_p(\textbf{f}_{i},\textbf{f}_{j})$ satisfies the properties of defined probabilistic proximity function.
\begin{itemize}
  \item[1.]  The range of the JS-divergence is within 0 and 1, thus $S_p(\textbf{f}_{i},\textbf{f}_{j})\in [0,1]$ is true.
  \item[2.]  Obviously, based on $\textbf{Eq (\ref{eq7})}$, it is easy to have $S_p(\textbf{f}_{i},\textbf{f}_{j})=\frac{1}{2}\left[d_{\text{KL}}(\bm{p}_i\|\Bar{\bm{p}}_{ij})+d_{\text{KL}}(\bm{p}_j\|\Bar{\bm{p}}_{ij}) \right]=S_p(\textbf{f}_{i},\textbf{f}_{j})$.
  \item[3.]  $S_p(\textbf{f}_{i},\textbf{f}_{j})\ge 0$, as a sum of nonnegative terms. To have $S_p(\textbf{f}_{i},\textbf{f}_{j})=0$, each term of $S_p(\textbf{f}_{i},\textbf{f}_{j})$ must be 0. Thus, $S_p(\textbf{f}_{i},\textbf{f}_{j})=0$ if and only if $d_{\text{KL}}(\bm{p}_i\|\Bar{\bm{p}}_{ij})=d_{\text{KL}}(\bm{p}_j\|\Bar{\bm{p}}_{ij})$. Since $d_{\text{KL}}(\bm{p}_i\|\bm{p}_j)=0$ if and only if $\bm{p}_i=\bm{p}_j$, thus, $S_p(\textbf{f}_{i},\textbf{f}_{j})=0$ if and only if $\textbf{f}_{i}=\textbf{f}_{j}$.
\end{itemize}
Now we prove property 4. Using the Shannon entropy, $H(\bm{p}_i)=-\sum_{p_i\in \bm{p}_i}p_i\log p_i$, the explicit form of $S_p(\textbf{f}_i,\textbf{f}_j)$ can be written as \begin{align}
    S_p(\textbf{f}_i,\textbf{f}_j)&=H(\Bar{\bm{p}}_{ij})-\frac{1}{2}\left[H(\bm{p}_i)+H(\bm{p}_j)\right] \nonumber
\end{align}
Assume that $H(\bm{\Bar{p}}_{ic})+H(\bm{\Bar{p}}_{cj})\ge H(\bm{\Bar{p}}_{ij})+H(\bm{\Bar{p}}_{c})$, thus, $S_p(\textbf{f}_i,\textbf{f}_c)+S_p(\textbf{f}_c,\textbf{f}_j)-S_p(\textbf{f}_i,\textbf{f}_j)$ can be rewritten as 
\begin{align}
    H(\bm{\Bar{p}}_{ic})-H(\bm{\Bar{p}}_{c})+H(\bm{\Bar{p}}_{cj})-H(\bm{\Bar{p}}_{ij}) \ge 0  \nonumber
\end{align}
Thus, $S_p(\textbf{f}_i,\textbf{f}_c)+S_p(\textbf{f}_c,\textbf{f}_j)\ge S_p(\textbf{f}_i,\textbf{f}_j)$ is true if and only if $H(\bm{\Bar{p}}_{ic})+H(\bm{\Bar{p}}_{cj})\ge H(\bm{\Bar{p}}_{ij})+H(\bm{\Bar{p}}_{c})$.
\end{proof}

\subsection{Learning Hybrid Proximity}
\subsubsection{Generalized Hybrid Metric Loss}
The learning objective loss function of the metric learning branch is the convex combination of geometric proximity loss and probabilistic proximity loss. As such, the objective is denoted by
\begin{align}
    \mathcal{L}^*_{\text{GHM}} = \lambda \mathcal{L}^*_{g} - (1-\lambda)\mathcal{L}^*_{p} \label{eq8}
\end{align}
where $\lambda \in [0,1]$ indicates the weighting factor to control the geometric proximity loss, $\mathcal{L}^*_{g}$, and the probabilistic proximity loss, $\mathcal{L}^*_{p}$. In this way, the network is able to capture both geometric and probabilistic information during the training process.

\subsubsection{Comparing With Geometric Proximity}
To show the advantage of including the probabilistic proximity loss in the metric learning branch using the probabilistic view, we compare the geometric proximity and the probabilistic proximity between two probability distribution.

Consider a KL-divergence between $\textbf{f}_i$ and $\textbf{f}_j$. For simplicity, we use $p(x)$ and $q(x)$ to represent $p(\textbf{f}_i)$ and $p(\textbf{f}_j)$, respectively, and assume $x\sim \mathcal{N}(\mu,\sigma^2)$. Thus, the expanded form of $d_{\text{KL}}(p\|q)$ for two Gaussians is denoted as
\begin{align}
    d_{\text{KL}}(p\|q)&=\int_{x}p(x) \log p(x) \ dx - \int_{x}p(x) \log q(x) \ dx  \label{eq9}
\end{align}
In here, we derive the result using the fact of \cite{robert1996intrinsic}. For the first term, $\int_{x}p(x) \log p(x) \ dx$ can be expanded as
\begin{align}
    &-\int_{x}p(x)\log\sqrt{2\pi\sigma^2_{p}} \ dx -  \int_{x}\frac{p(x)(x-\mu_{p})^2}{2\sigma^2_{p}} \ dx \nonumber \\
    &= -\log\sqrt{2\pi\sigma^2_{p}} - \frac{1}{2\sigma^2_{p}}\int_{x}p(x)(x-\mu_{p})^2 \ dx  \label{eq10}
\end{align}
Next, we expand the quadratic form:
\begin{align}
    &-\log\sqrt{2\pi\sigma^2_{p}}- \frac{1}{2\sigma^2_{p}}\left[\EX_p(x^2)-\EX_p(x)^2 \right] \nonumber \\
    &=-\log\sqrt{2\pi\sigma^2_{p}}-\frac{1}{2}  \label{11}
\end{align}
Following the same derivation, $\int_{x}p(x) \log q(x) \ dx$ can be expand by
\begin{align}
    \int_{x}p(x) \log q(x) \ dx =-\log\sqrt{2\pi\sigma^2_{q}}-\frac{\left[\sigma^2_{q}+(\mu_p-\mu_q)^2\right]}{2\sigma^2_{q}}  \label{eq12}
\end{align}
Assume $\sigma^2_{p}=\sigma^2_{q}=c$, $c$ is a constant, based on \textbf{Eq (\ref{eq10})-(\ref{eq12})}, the KL-divergence between $p(x)$ and $q(x)$ is given by
\begin{align}
    d_{\text{KL}}(p\|q) = -\frac{1}{2}-\frac{1}{2c}+\frac{1}{2c}\left(\mu_p-\mu_q\right)^2 \label{eq13}
\end{align}
that is a linear function consisting of $L_2$ distance between two mean values, showing that the probabilistic proximity also considers the variation of the sampling distribution, while the geometric proximity does not. This derivation also supports the phenomenon in \textbf{Figure \ref{f1}}.

\subsection{Network Implementation Details}
As illustrated in \textbf{Figure \ref{framework}}, the proposed framework consists of a feature extraction backbone and two supervised learning branches: one for metric learning and one for classification. We used a pretrained ResNet18 \cite{he2016deep}, following the same setting as the previous work \cite{li2021rotation}. We used max pooling on the attention map after the last layer of the residual block in ResNet18. Then, we flatted the output to a vector and sequentially connect it with a MLP layer, batch normalization, and ReLU to reduce the feature dimension to 128. Next, each $\textbf{f}_i$ 1) was connected with a $L_2$ normalization layer, i.e., $\|\textbf{f}_i\|=1$ to calculate the hybrid proximity of the metric learning branch, and 2) connect to another MLP layer and Softmax for classification. 

The classification branch is to take the input batch $\{x\}_{i=1}^{b}$ to generate a prediction output. We optimized the cross-entropy loss, $ \mathcal{L}_{\text{CE}}$, together with the metric learning loss, $ \mathcal{L}^*_{\text{GHM}}$, in a multi-task learning scheme. Thus, we defined our total objective loss as the weighted combination of a metric learning branch and a classification branch. The learning objective loss is denoted by 
\begin{align}
    \mathcal{L}_{\text{total}}=\beta \mathcal{L}^*_{\text{GHM}}+\mathcal{L}_{\text{CE}}
    \label{eq14}
\end{align}
where $\beta$ indicates the weighting factor to control the importance of the GHM-Loss In our experiments, we set $\beta=1$ and $\lambda=0.5$, we also analyze the effects of both $\beta$ and $\lambda$ using a grid search. Each input image of a batch was randomly scaled within a factor range of [0.3, 1.0], and cropped into patches of size 224 x 224. We set the batch size $b=8$ in the experiment and trained our framework using an Adam optimization, the learning rate and weight decay are set to 0.0001. We train our network for 2000 epochs. The whole framework was implemented using python 3.8, Scikit-Learn 0.24.1, Pytorch 1.9.1, and Cuda 11.1 with a NVIDIA GeForce GTX 1660 SUPER GPU.

\section{Data and Experiments}
\subsection{Datasets}
 To show the effectiveness of our method, same as Li et al \cite{li2021rotation}, we perform two binary (normal and abnormal) classification tasks by diagnosing pathological myopia (PM) and age-related macular degeneration (AMD) on two public ophthalmic disease datasets of iChallenge-PM and iChallenge-AMD. 
 
 \subsubsection{iChallenge-PM}
 iChallenge-PM \cite{fu2019palm} contains 1200 annotated retinal fundus images in which 50\% are PM subjects. More details of the iChallenge-PM dataset can be found on the \cite{fu2019palm}. We perform a 10-fold cross-validation to evaluate our method. 
 
 \subsubsection{iChallenge-AMD}
There is a total of 1200 color fundus images of the iChallenge-AMD dataset \cite{fang2022adam}, in which 77\% are non-AMD subjects and 23\% are AMD subjects. It provides the disc boundaries and fovea locations, as well as the boundaries of kinds of lesions. More details of the iChallenge-AMD dataset can be found on \cite{fang2022adam}. Note, that we only used the training dataset (400 fundus images) since only the training dataset is released with annotations. We perform a 10-fold cross-validation to evaluate our method. 

\subsection{Model Comparison Setting}
\subsubsection{Evaluation Metrics}
We used AUC, accuracy, precision, recall, and F1-score to assess the classification performance. AUC stands for Area Under the Receiver Operating Characteristic (ROC) curve. The definition of accuracy, precision, recall, and F1-score are denoted by:
\begin{align}
    \text{Accuracy}&=(TP+TN)/(TP+TN+FP+FN)\nonumber \\
    \text{Precision}&=TP/(TP+FP)\nonumber \\
    \text{Recall}&=TP/(TP+FN)\nonumber \\
    \text{F1}&=2*(\text{Precision}*\text{Recall})/({\text{Precision}+\text{Recall}}) \nonumber
\end{align}
where TP, TN, FP, and FN indicate the true positive, true negative, false positive, and false negative, respectively. 

To provide the statistical analysis of our method, we conducted a non-parametric Wilcoxon test \cite{woolson2007wilcoxon} with a $\alpha$ level of 0.05. A p-value less than 0.05 is considered as statistical significant for all inference. All statistical tests in the experiments were
performed using R-4.0.3 (RStudio, Boston, MA, USA).

\subsubsection{Competing State-of-the-Art Methods}
To have a fair comparison, we trained all peer methods with the pretrained ResNet18 with the same hyperparameters, network architectures, and optimizer under the 10-fold cross-validation. Since our framework consists of metric learning and classification branches, we fix the classification branch and only modify the metric learning part when compared with other metric learning methods in the experiment. Our proposed method was compared with other deep metric learning methods, Siamese \cite{koch2015siamese}, Triplet \cite{hoffer2015deep}, SupCon \cite{khosla2020supervised},  N-pair \cite{sohn2016improved}, and InfoNCE \cite{oord2018representation}.  We run these metric learning methods with the code released on iChallenge-PM and iChallenge-AMD datasets. We also provided a supervised ‘Baseline’ method by modifying the output layer of the last fully connected layer of the ResNet18 to 2 and trained with cross-entropy loss.

\subsection{Comparison on the iChallenge-PM Dataset}
We compared with other state-of-the-art methods on the iChallenge-PM Dataset. The results are shown in $\textbf{Table \ref{Tab1}}$. We found that each method can achieve over 95\% prediction performance on all evaluation metrics, which indicates that the patterns of pathological myopia in color fundus images are obvious. We can see that N-pair \cite{sohn2016improved} achieved a limited result and is due to this method requires large, annotated training data that may not be suitable for the color fundus images. Notably, our method significantly outperformed other peer metric learning methods with 99.08\% (p$<$0.0001)  on AUC and 99.01\% (p$<$0.0001)  on accuracy for PM diagnosis. These results further demonstrate the effectiveness of our method compared to other state-of-the-art metric learning methods. 
\begin{table}[ht]
    \centering
    \caption{Model Comparisons with other deep metric learning methods on the iChallenge-PM Dataset (UNIT: \%).}
    \begin{tabular}{c|ccccc}
     \hline
     	& AUC & Accuracy & Precision & Recall & F1\\
     \hline
     Baseline & 96.01 & 95.45 &	94.51 &	97.25 &	95.34 \\
     \hline
     Siamese \cite{koch2015siamese} & 97.45 & 97.30 & 96.15 & 96.60 & 96.58	\\
     Triplet \cite{hoffer2015deep} & 97.95 & 98.64 & 97.49 & 96.14 & 97.21  \\
     SupCon  \cite{khosla2020supervised} & 98.06 & 98.22 & \textbf{98.36} & 97.29 & 97.64  \\
     N-pair  \cite{sohn2016improved} & 95.36 & 95.83 & 96.41 & 97.25 & 96.12\\
     InfoNCE  \cite{oord2018representation} & 98.11 & 97.91 & 96.83 & 97.59 & 97.36  \\
     \textbf{Ours} & \textbf{99.08} & \textbf{99.01} & 98.08 & \textbf{99.12} & \textbf{98.40} \\
     \hline
    \end{tabular}
    \label{Tab1}
\end{table}

\subsection{Comparison on the iChallenge-AMD Dataset}
We compared with other state-of-the-art methods on the iChallenge-AMD Dataset. As shown in \textbf{Table \ref{Tab2}}, we can see that our method achieved the best prediction performance among other competing metric learning methods. Compared to the second-best method, InfoNCE \cite{oord2018representation}, our method significantly improved the performance, i.e., 78.69\% vs. 76.75\% (p$<$0.0001) on AUC and 88.04 \% vs. 86.51\% (p$<$0.0001) on accuracy. Notably, our method also outperformed the supervised ‘Baseline’ method on all evaluation metrics. These results demonstrated the effectiveness of the proposed method.
\begin{table}[ht]
    \centering
    \caption{Model Comparisons with other deep metric learning methods on the iChallenge-AMD Dataset (UNIT: \%).}
    \begin{tabular}{c|ccccc}
     \hline
     	& AUC & Accuracy & Precision & Recall & F1\\
     \hline
     Baseline & 76.51 & 84.16 &	82.54 &	76.18 &	78.86 \\
     \hline
     Siamese \cite{koch2015siamese}  & 67.58 &	82.45 &	72.54 &	68.26 &	70.14 \\
     Triplet \cite{hoffer2015deep} & 69.52 & 84.29 & 76.87 & 72.48 & 73.21 \\
     SupCon  \cite{khosla2020supervised} & 73.24 & 85.64 & 78.42 & 74.15 & 76.05 \\
     N-pair  \cite{sohn2016improved} & 69.58 & 83.41 & 75.14 & 70.54 &71.86	 \\
     InfoNCE  \cite{oord2018representation} & 76.75 & 86.51 & \textbf{85.36} & 72.35 & 77.95	 \\
     \textbf{Ours} & \textbf{78.69} & \textbf{88.04} & 82.95 & \textbf{75.28} & \textbf{78.24} \\
     \hline
    \end{tabular}
    \label{Tab2}
\end{table}

\subsection{Comparison with Transfer Learning Models}
To show the robustness of learned features of our method, we compared our method with the ImageNet pretrained models, including VGG-19 \cite{simonyan2014very}, InceptionNet v1 \cite{szegedy2015going}, and EfficientNet B0 \cite{tan2019efficientnet} on the iChallenge-AMD dataset. We modified the output channel of the last fully connected layer in each pretrained model to 2 and trained them with cross-entropy loss. To have a fair comparison, all the models were trained with the same number of epochs, learning rate, and weight decay term on a 10-fold cross validation. The results are shown in $\textbf{Table \ref{Tab3}}$. We can see that Efficient Net achieves the best prediction performance among the transfer learning models. Compared to Efficient Net,  it is observed that our method can achieve a higher prediction performance with around 1.5\% (p$<$0.0001) on AUC and 7\% (p$<$0.0001) on accuracy. Note, we trained our method with only 400 color fundus images and performed better than ImageNet models, which were pretrained with more than 1 million natural images. With this observation, the results further show the practical value of our method. 
\begin{table}[ht]
    \centering
    \caption{Model Comparisons with ImageNet Transfer Learning Models on the iChallenge-AMD Dataset (UNIT: \%).}
    \begin{tabular}{c|ccccc}
     \hline
     	& AUC & Accuracy & Precision & Recall & F1\\
     \hline
     VGG-19 \cite{simonyan2014very} & 74.14 & 81.52 & 76.54 & 72.36 & 73.89 \\
     Inception v1 \cite{szegedy2015going} & 76.32 & 77.35 & 78.39 & 75.54 & 76.28\\
     Efficient B0  \cite{tan2019efficientnet} & 77.25 & 81.52 & 80.32 & 79.25 & 79.52\\
     \textbf{Ours} & \textbf{78.69} & \textbf{88.04} & \textbf{82.95} & \textbf{75.28} & \textbf{78.24} \\
     \hline
    \end{tabular}
    \label{Tab3}
\end{table}

\subsection{Analytical Study}
\subsubsection{Importance of the GHM-Loss}
\begin{table}[ht]
    \centering
    \caption{The importance of the GHM-Loss in the metric learning branch on the iChallenge-AMD Dataset (UNIT: \%).}
    \begin{tabular}{c|ccccc}
     \hline
     	& AUC & Accuracy & Precision & Recall & F1\\
     \hline
     $\beta=0.0$ & 75.41 & 83.21 & 80.54 & 72.88 & 76.15 \\
     $\beta=0.5$ & 76.85 & 85.42 & 80.95 & 74.54 & 77.28\\
     $\beta=1.0$ & \textbf{78.69} & \textbf{88.04} & \textbf{82.95} & \textbf{75.28} & \textbf{78.24} \\
     $\beta=2.0$ & 72.45 & 79.41 & 77.66 & 70.23 & 73.59\\
     \hline
    \end{tabular}
    \label{Tab4}
\end{table}

The proposed method consists of metric learning branch and classification branch in a multi-task scheme, in which we trained GHM-Loss together with the cross-entropy loss. In this section, we analyzed the importance of the GHM-Loss of our method on the iChallenge-AMD dataset. We first fix the $\lambda =0.5$ in GHM-Loss and trained our framework with different $\beta$ in $\textbf{Eq}$ (\ref{eq14}), where $\beta$ is the importance of the metric learning branch. $\beta=0.0$ denotes that the framework is only trained with the cross-entropy loss. As $\beta$ increases, the more weight or importance of the GHM-Loss in the network training. 

The results are shown in $\textbf{Table \ref{Tab4}}$. As we can see, when $\beta=0.0$, the network only learns the classification branch and the result is 75.41\% on AUC and 83.21\% on accuracy. As $\beta$ increases, we found that the prediction performance improves to the best when $\beta $ reached 1 (e.g., 78.69\% on AUC, 88.04\% on accuracy). However, when $\beta$ continues increasing, the prediction performance starts to drop apparently from  78.69\% to 72.45\% on AUC. The comparison shows that both the metric learning branch and classification branch equally contributed to our framework for PM diagnosis.

\subsubsection{Effects of Weighting Factors in the GHM-Loss}
\begin{figure}[ht]
    \centering
    \includegraphics[width=8.8cm]{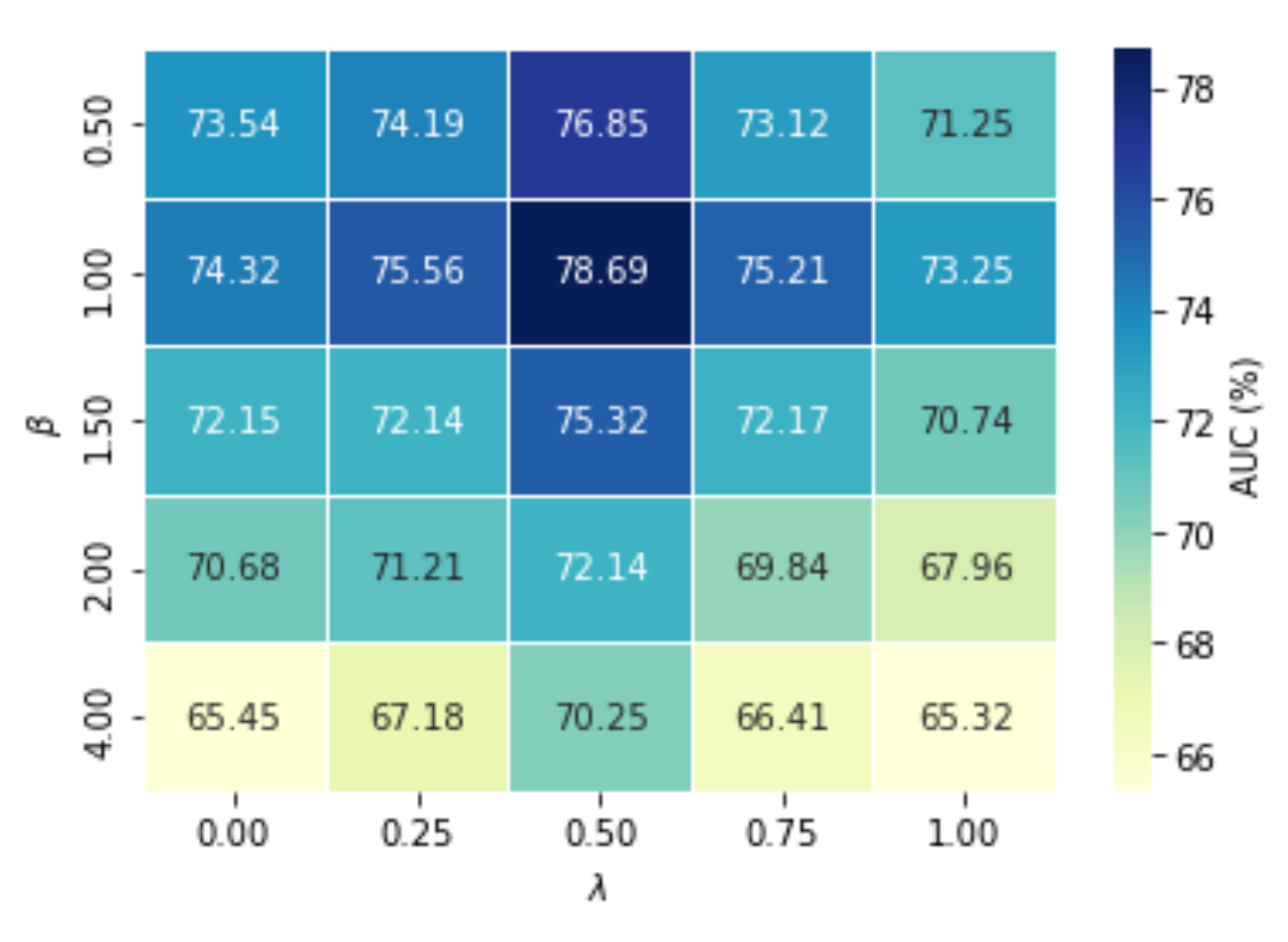}
    \caption{Classification performance comparison on the iChallenge-AMD Dataset with different weighting factors $\beta$ and $\lambda$ of the GHM-Loss. We use AUC to choose the optimal $\beta$ and $\lambda$ using a grid search.}
    \label{heapmap}
\end{figure}

We analyzed the effects of the weighting factors, i.e., $\beta$, $\lambda$ in the GHM-Loss on the iChallenge-AMD Dataset, in which $\beta$ indicates the importance of the metric learning branch of our method and $\lambda$ controls the weight size between geometric proximity and probabilistic proximity of the GHM-Loss. Note, $\lambda=0.0$ denotes that only probabilistic proximity was considered between $\textbf{f}_i$ and $\textbf{f}_j$. As we can see in \textbf{Figure \ref{heapmap}}, for each fix $\beta$, the classification performance increases to the best performance when $\lambda$ is reached 0.5 and drops apparently as it continues to increase. These results demonstrate that 1) both metric learning and classification branches are useful of our method and 2) both geometric and probabilistic proximity should be captured between $\textbf{f}_i$ and $\textbf{f}_j$ in the training. 
\subsubsection{Visualization of the Feature Distribution}
\begin{figure}[ht]
    \centering
    \includegraphics[width=8.8cm]{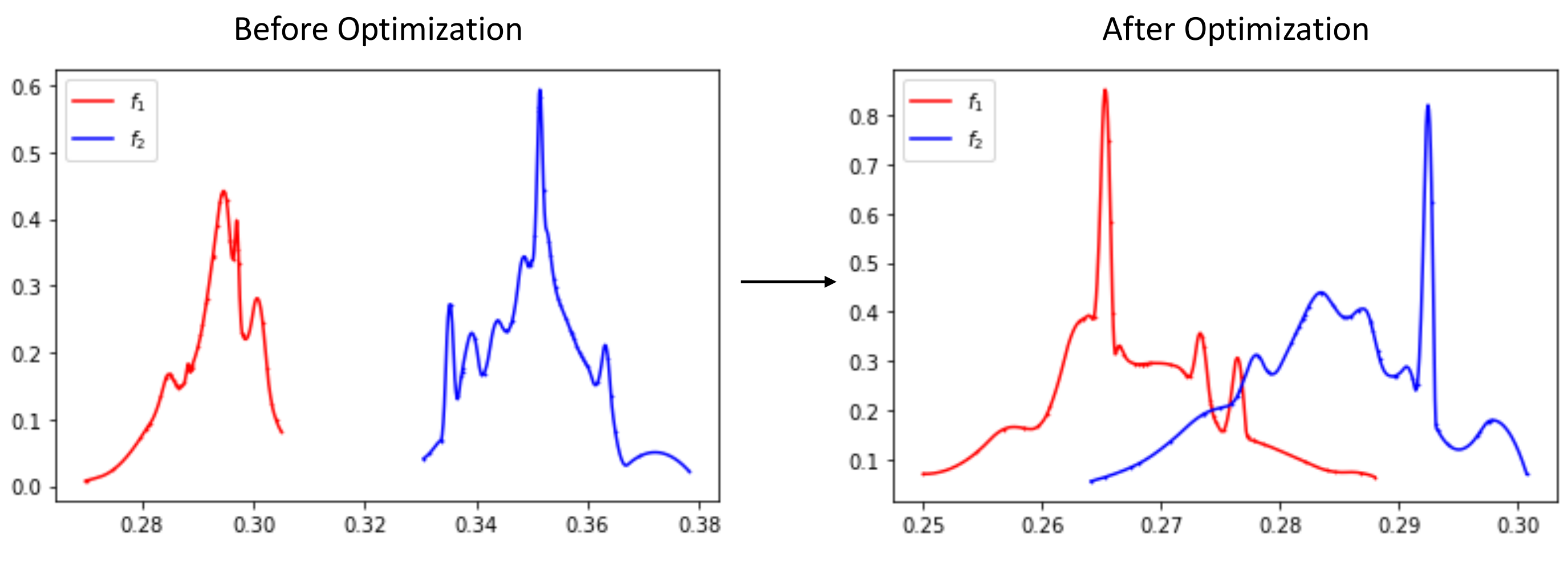}
      \caption{The feature distribution between a positive pair of $\textbf{f}_1$ (red line) and $\textbf{f}_2$ (blue line) of color fundus images during the training process on the iChallenge AMD dataset. We applied Gaussian mixture model (GMM) to approximate the empirically distribution of these features. The probabilistic proximity between $\textbf{f}_1$ (red line) and $\textbf{f}_2$ are  reduced after optimization.}
    \label{dist}
\end{figure}

We visualized the feature embedding distribution, i.e., $\textbf{f}_1$ (red line) and $\textbf{f}_2$, after ResNet18 for a positive pair color fundus image on the iChallenge AMD dataset. The feature distributions are shown in \textbf{Figure \ref{dist}}. Before optimization, we can see that the distribution of feature embeddings from a positive pair sample are independent without overlaps. However, the probabilistic distance of $\textbf{f}_1$ (red line) and $\textbf{f}_2$ is reduced and stays close to each other after optimizing the network. Since we use GMM to approximate the empirical distribution of each feature embedding, the probability parameters of $\mu$ and $\sigma$ of all the images with the same label should be closed to each other, thus, resulting in the similar probability densities. This visualization also demonstrate that the proposed GHM-Loss can efficiently capture the probabilistic patterns during the training process.

\section{Discussion}
Metric learning is an important technique in visual representation area by learning the distance metric, which can be further used to perform supervised and unsupervised learning tasks, such as image classification \cite{khosla2020supervised,chen2020simple,he2020momentum}, image clustering \cite{do2021clustering,li2021contrastive}, and object detection \cite{kim2022spatial,xie2021detco}, etc. With the advances of deep learning techniques, deep metric learning has been widely studied in the metric learning research community. Although promising results were obtained on previous works \cite{koch2015siamese,hoffer2015deep,sohn2016improved,khosla2020supervised,oord2018representation}, these methods usually ignore the probability distribution of the feature embeddings during the training process, which may lead an inaccurate prediction. In this work, we present a novel supervised metric learning method that consists of learning both geometric and probabilistic proximity for image recognition. We formulate a \textbf{\emph{Generalized Hybrid Metric Loss}} (GHM-Loss) to better learn the distance representation, where geometric-based distance and probabilistic-based distance are learned. Our method is validated on two public ophthalmic disease datasets (e.g., iChallenge-PM and iChallenge-AMD), in which our method can significantly outperform other state-of-the-art metric learning methods. With a convex combination of the geometric proximity and probabilistic proximity, our method consistently achieves the best prediction performance than the individual proximity.

Although our method outperforms other state-of-the-art metric learning methods, it comes with limitations. Our method is a supervised learning approach, which relies on a large number of annotated training data, and it is costly to obtain. In future, we will investigate the unsupervised metric learning approach or self-supervised learning approach to address the human effort issue on image recognition communities. The exploration of probabilistic unsupervised/self-supervised metric learning would be our future work.

\section{Conclusion}
In this paper, we present a novel supervised metric learning method for image recognition. Our main idea is to learn a hybrid proximity that consists of both geometric-based metric and probabilistic-based metric. The geometric proximity of proposed GHM-Loss helps the model learn the similarity information under the Euclidean space and the probabilistic proximity proposed GHM-Loss learns the similarity under the empirical probability distribution. With extensive experiments, our method consistently achieves the excellent prediction performance compared with the other state-of-the-art metric learning methods, showing the effectiveness of learned distance features of our method in image recognition.

\bibliographystyle{IEEEtran}
\bibliography{ref.bib}
\end{document}